\documentclass{article}

\PassOptionsToPackage{numbers, compress}{natbib}

\usepackage{graphicx}

\usepackage[preprint]{neurips_2022}

\usepackage[utf8]{inputenc} 
\usepackage[T1]{fontenc}    
\usepackage{hyperref}       
\usepackage{url}            
\usepackage{booktabs}       
\usepackage{amsfonts}       
\usepackage{nicefrac}       
\usepackage{microtype}      
\usepackage{xcolor}         

\title{ColBERT: Using BERT Sentence Embedding in Parallel Neural Networks for Computational Humor}

\author{%
  Issa Annamoradnejad\thanks{Corresponding author: Annamoradnejad is with the Department of
Computer Engineering, Sharif University of Technology, Tehran, Iran}\\
  \texttt{i.moradnejad@gmail.com} \\
   \And
   Gohar Zoghi \\
   \texttt{zoghi.g@goums.ac.ir} \\
}

\begin{document}

\maketitle

\begin{abstract}
  Automation of humor detection and rating has interesting use cases in modern technologies, such as humanoid robots, chatbots, and virtual assistants. In this paper, we propose a novel approach for detecting and rating humor in short texts based on a popular linguistic theory of humor. The proposed technical method initiates by separating sentences of the given text and utilizing the BERT model to generate embeddings for each one. The embeddings are fed to separate lines of hidden layers in a neural network (one line for each sentence) to extract latent features. At last, the parallel lines are concatenated to determine the congruity and other relationships between the sentences and predict the target value. We accompany the paper with a novel dataset for humor detection consisting of 200,000 formal short texts. In addition to evaluating our work on the novel dataset, we participated in a live machine learning competition focused on rating humor in Spanish tweets. The proposed model obtained F1 scores of 0.982 and 0.869 in the humor detection experiments which outperform general and state-of-the-art models. The evaluation performed on two contrasting settings confirm the strength and robustness of the model and suggests two important factors in achieving high accuracy in the current task: 1) usage of sentence embeddings and 2) utilizing the linguistic structure of humor in designing the proposed model.
\end{abstract}

\section{Introduction}
\label{sec:intro}

In Interstellar (2014 movie), a future earth is depicted where robots easily understand and use humor in their connections with their owners and humans can set the level of humor in their personal robots\footnote{Tarzs, in the movie.}. While we may have a long road toward the astral travels, we are very close in reaching high-quality systems injected with adjustable humor. 

Humor, as a potential cause of laughter, is an important part of human communication, which not only makes people feel comfortable but also creates a cozier environment \cite{castro2016joke}. Automatic humor detection in texts has interesting use cases in building human-centered artificial intelligence systems such as humanoid robots, chatbots, and virtual assistants. An appealing use case is to identify whether an input command should be taken seriously or not, which is a critical step to understanding the real motives of users, returning appropriate answers, and enhancing the overall experience of users with the AI system. A more advanced outcome would be the injection of humor into computer-generated responses, thus making the human-computer interaction more engaging and interesting \cite{niculescu2013making}. This is an outcome that is achievable by setting the level of humor in possible answers to the desired level, similar to the mentioned movie.

Humor can be attained through several linguistic or semantic mechanisms, such as wordplay, exaggeration, misunderstanding, and stereotype. Researchers proposed several theories to explain humor functionality as a trait, one of which is called "incongruity theory", where laughter is the result of realizing incongruity in the narrative. A general version states that a common joke consists of a few sentences that conclude with a punchline. The punchline is responsible for bringing contradiction into the story, thus making the whole text laughable. In other words, any sentence can be non-humorous in itself, but when we try to comprehend all sentences together in one context or in a single line of story, the text becomes humorous. 

By focusing on this widely used structure of jokes, we believe and show that it is required to view and encode each sentence separately and capture the underlying relation between sentences in a proper way. As a result, our proposed model for the task of humor detection is based on creating parallel paths of neural network hidden layers, in addition to encoding a given text as a whole.

In short, the proposed approach initiates by separating the text into its sentences. Then, it utilizes the BERT model in order to encode each sentence and the whole text into embeddings. Next, the embeddings will be fed into parallel hidden layers of a neural network to extract latent features regarding each sentence. The last three layers combine the output of all previous lines of hidden layers to determine the relationship between the sentences to predict the final output. In theory, these final layers are responsible for determining the congruity or detecting the transformation of the reader's viewpoint after reading the punchline.

In addition to proposing a model for humor detection, we curated a large dataset for the binary task of humor detection. Previous attempts combined formal non-humorous texts with informal humorous short texts, which due to the incompatible statistics of the parts (text length, words count, etc.) makes it more likely to detect humor with simple analytical models and without understanding the underlying latent lingual features and structures. To address this problem, we applied multiple analytical pre-processing steps to trim outlier texts which resulted in a dataset that is statistically similar based on the target class.

To test the robustness and stability of the proposed model, we evaluate its performance in two different settings: 
\begin{enumerate}
    \item Evaluation on the novel dataset (short formal English texts): The model is evaluated for the binary task of humor detection on the novel dataset, and 
    \item Evaluation in a live machine-learning competition (Spanish informal texts): The model competes against strong real teams to detect and rate humor in informal Spanish texts (variant lengths).
\end{enumerate}

We summarize our contributions as follows:
\begin{itemize}
    \item We propose an automated approach for humor detection in texts that is based on a general theory of humor. We introduce the model architecture and components in detail.
    \item We introduce a new dataset for the task of humor detection, entitled the “ColBERT dataset", which contains 200k short texts (100k positive and 100k negative). We reduced or completely removed issues prevalent in the existing datasets to build a proper dataset for the task.
    \item We evaluate the performance of our proposed model on the novel dataset in comparison with five strong baselines.
    \item We further evaluate its accuracy and robustness in a data science competition for Spanish texts.
\end{itemize}

The structure of this article is as follows: Section 2 reviews past works on the task of humor detection with a focus on transfer learning methods. Section 3 describes the data collection and preparation techniques and introduces the new dataset. Section 4 elaborates on the methodology, and section 5 presents our experimental results. Section 6 is the concluding remarks.

\section{Literature Review}

With advances in NLP, researchers applied and evaluated state-of-the-art methods for the task of humor detection. This includes using statistical and N-gram analysis \cite{taylor2004computationally}, Regression Trees \cite{purandare2006humor}, Word2Vec combined with K-NN Human Centric Features \cite{yang2015humor}, and Convolutional Neural Networks \cite{chen2018humor} \cite{weller2019humor}. 

Weller and Seppi \cite{weller2019humor} focused on the task of humor detection by using a Transformer architecture. The work approached the task by learning on ratings taken from the popular Reddit r/Jokes thread (13884 negative and 2025 positives). 
Kramer \cite{kramer2021laugh} suggest that our sense of humour is acutely aware of our flaws and tackle the problem in words of error detection.

There are emerging tasks related to humor detection. Ref \cite{yang2019predicting} focused on predicting humor by using audio information, hence reached 0.750 AUC by using only audio data. A good number of research is focused on the detecting humor in non-English texts, such as on Spanish \cite{chiruzzo2019overview, ismailov2019humor, giudice2019aspie96}, Chinese \cite{yang2019predicting}, and English-Hindi \cite{khandelwal2018humor}.

With the popularity of transfer learning, some researchers focused on using pre-trained models for several tasks of text classification. 
Among them, BERT \cite{devlin2018bert} utilizes a multi-layer bidirectional transformer encoder consisting of several encoders stacked together, which can learn deep bi-directional representations. Similar to previous transfer learning methods, it is pre-trained on unlabeled data to be later fine-tuned for a variety of tasks. It initially came with two model sizes (BERT\textsubscript{BASE} and BERT\textsubscript{LARGE}) and obtained eleven new state-of-the-art results. Since then, it was pre-trained and fine-tuned for several tasks and languages, and several BERT-based architectures and model sizes have been introduced (such as Multilingual BERT, RoBERTa \cite{liu2019roberta}, ALBERT \cite{lan2019albert} and VideoBERT \cite{sun2019videobert}).

\section{Data}

Existing humor detection datasets use a combination of formal texts and informal jokes with incompatible statistics (text length, words count, etc.), making it more likely to detect humor with simple analytical models and without understanding the underlying latent connections. Moreover, they are relatively small for the tasks of text classification, making them prone to over-fit models. These problems encouraged us to create a new dataset exclusively for the task of humor detection, where simple feature-based models will not be able to predict without an insight into the linguistic features.

In this section, we will introduce data collection method, data sources, filtering methods, and some general statistics on the new dataset.

\begin{table}
  \caption{Datasets for the binary task of humor classification}
  \label{table-1}
  \centering
  \begin{tabular}{l|ll}
    \hline
    \multicolumn{2}{r}{Parts}                   \\ \hline
     Dataset     &   \#Positive  &  \#Negative \\ \hline
    16000 One-Liners \cite{mihalcea2005making} & 16,000   &	16,002     \\
    Pun of the Day \cite{yang2015humor}     & 2,423  &	2,403      \\
    PTT Jokes \cite{chen2018humor}     &     1,425	&   2,551  \\
    English-Hindi \cite{khandelwal2018humor}     & 1,755	&   1,698  \\ \hline
    ColBERT     &       100,000	&   100,000  \\ \hline
  \end{tabular}
\end{table}

\begin{table*}[t]
  \caption{General statistics of the ColBERT dataset (100k positive, 100k negative)}
  \label{table-2}
  \centering
  \begin{tabular}{p{1.1cm}|p{0.9cm}p{0.95cm}p{1.2cm}p{2.1cm}p{1.8cm}p{1.6cm}p{2.2cm}}
\hline & \#chars & \#words & \#unique words & \#punctuation & \#duplicate words & sentiment polarity & sentiment subjectivity \\ \hline
mean & 71.561  & 12.811  & 12.371         & 2.378          & 0.440    & 0.051              & 0.317                  \\
std  & 12.305  & 2.307   & 2.134          & 1.941          & 0.794     & 0.288              & 0.327                  \\
min  & 36      & 10      & 3              & 0              & 0       & -1.000             & 0.000                  \\
median & 71      & 12      & 12             & 2              & 0       & 0.000              & 0.268                  \\
max  & 99      & 22      & 22             & 37             & 13       & 1.000              & 1.000     \\ \hline
  \end{tabular}
\end{table*}

\subsection{Data Collection}
\label{sec:datacol}

We carefully analyzed existing datasets (exclusively on news stories, news headlines, Wikipedia pages, tweets, proverbs, and jokes) with regard to table size, character length, word count, and formality of language. Table~\ref{table-1} present an overview of the existing humor detection datasets (binary task) and highlight their size and data sources. There are other datasets focused on closely related tasks, e.g. the tasks of punchline detection and success (whether or not a punchline triggers laughter) \cite{chen2017predicting,hasan2019ur}, or on using speak audio and video to detect humor \cite{bertero2016deep,hasan2019ur}. 

To curate the new dataset, we chose two data sources with formal texts (one with humor texts and one without) that were syntactically similar in our initial analysis.

\begin{enumerate}
    \item News category dataset \cite{dataset2018news}, published under CC0 Public Domain, consist of 200k Huffington Post news headlines from 2012-2018 and contains headlines, corresponding URLs, categories and full stories. The stories are scattered in several news categories, including politics, wellness, entertainment and parenting.

    \item Jokes dataset contains 231,657 jokes/humor short texts, crawled from Reddit communities\footnote{Mostly from /r/jokes and /r/cleanjokes subreddits.}. The dataset is compiled as a single \verb+csv+ file with no additional information about each text (such as the source, date, etc) and is available at Kaggle. Ref \cite{chen2018humor} combined this dataset with the WMT162 English news crawl, but did not publicly publish the dataset. Ref \cite{weller2019humor} also combined this dataset with extracted sentences from the WMT162 news crawl and made it publicly available.
\end{enumerate}

Next, we performed a few preprocessing steps to create a dataset that is syntactically the same for both target class.

\subsection{Preprocessing and Filtering}

This part contains a few preprocessing steps to create the new dataset.

The initial step was to drop duplicate texts, as we identified duplicate rows in both datasets. Dropping duplicate rows removed 1369 rows from the jokes dataset and 1558 rows from the news dataset.

Then, to make the lexical statistics similar, we calculated the average and standard deviation of number of characters and words for each group. Then, we started selecting texts in pairs in a way that each text with some statistic will have a similar text from the other group. As a result, we only kept texts with character length between 30 and 100, and word length between 10 and 18. Resulting data parts have very similar distribution with regard to these statistics.

In addition, we noticed that headlines in the news dataset use Title Case\footnote{All words are capitalized, except non-initial articles like “a, the, and”, etc.} formatting, which was not the case with the jokes dataset. Thus, we decided to apply Sentence Case\footnote{Capitalization as in a standard English sentence, e.g., “Witchcraft is real.”.} formatting to all news headlines by keeping the first character of the sentences in capital and lower-casing the rest. This simple modification helps to prevent simple classifiers from reaching perfect accuracy.

Finally, we randomly selected 100k rows from both datasets and merged them together to create an evenly distributed dataset.

\begin{table}
  \caption{A few examples from the novel dataset}
  \label{table-exa}
  \centering
  \begin{tabular}{p{8cm}|l}
  \hline
Text & Is humor?    \\ \hline
Why your purse is giving you back pain... and 11 ways to fix it & False \\
Why was the fruit/vegetable hybrid upset? he was a melon-cauliflower. & True \\
On set with Paul Mitchell: from our network & False \\
Starting a cover band called a book so no one can judge us. & True \\

\hline
  \end{tabular}
\end{table}

\subsection{Dataset Statistics}

Dataset\footnote{The dataset is available at: https://github.com/Moradnejad/ColBERT-Using-BERT-Sentence-Embedding-for-Humor-Detection} contains 200k labeled short texts equally distributed between humor and non-humor. It is much larger than the previous datasets (Table~\ref{table-1}) and it includes texts with similar textual features. Table~\ref{table-exa} contains a few random examples from the dataset and Table~\ref{table-2} displays general textual statistics of the dataset. The correlation between character count and the target is insignificant. 

The sentiment polarity and subjectivity are calculated for all texts using \verb+TextBlob+ python library. The correlation analysis resulted in coefficients of -0.09 and +0.02 for polarity and subjectivity, respectively, which suggest no notable connection between the target value and sentiment features.

\section{Proposed Method}
In this section, we will explore our proposed method for the task of humor detection. From a technical viewpoint, we are proposing a supervised binary classifier that takes a string as input and determines if the given text is humorous or not.

\subsection{Humor Structure}
First, we take a look at the general structure of a joke to understand the underlying linguistic features that makes a text laughable. 

There has been a long line of works in linguistics of humor that classify jokes into various categories based on their structure or content. Many suggested that humor arises from the sudden transformation of an expectation into nothing \cite{kant1913kritik}. In this way, punchline, as the last part of a joke, destroys the perceiver's previous expectations and brings humor to its incongruity. Based on this, some popular theories suggest that the structure of a joke involve two or three stages of storytelling that conclude with a punchline \cite{eysenck1942appreciation, suls1972two}. 

Raskin \cite{raskin2012semantic} presented Semantic Script Theory of Humor (SSTH), a detailed formal semantic theory of humor. The SSTH has the necessary condition that a text has to have two distinct related scripts that are opposite in nature, such as real/unreal, possible/impossible. For example, let us review a typical joke:
    \begin{quote}
        “Is the doctor at home?” the patient asked in his bronchial whisper. “No,” the doctor’s young and pretty wife whispered in reply. “Come right in.” \cite{raskin2012semantic}
    \end{quote}

This is compatible with the two-staged theory which ends with a punchline. The punchline is related to previous sentences but is included as opposition to previous lines in order to transform the reader's expectation of the context.

\subsection{Model Architecture}
Based on the presented short introduction to the structure of humor, if one reads sentences of a joke separately, they are most likely to be found as normal and non-humorous texts. On the other hand, if we try to comprehend all sentences together in one context or in one line of story, the text becomes humorous. Our proposed method utilizes this linguistic characteristic of humor in order to view or encode sentences separately and extract mid-level features using hidden layers.

\begin{figure*}[t]
\begin {center}
 \includegraphics[scale=0.8]{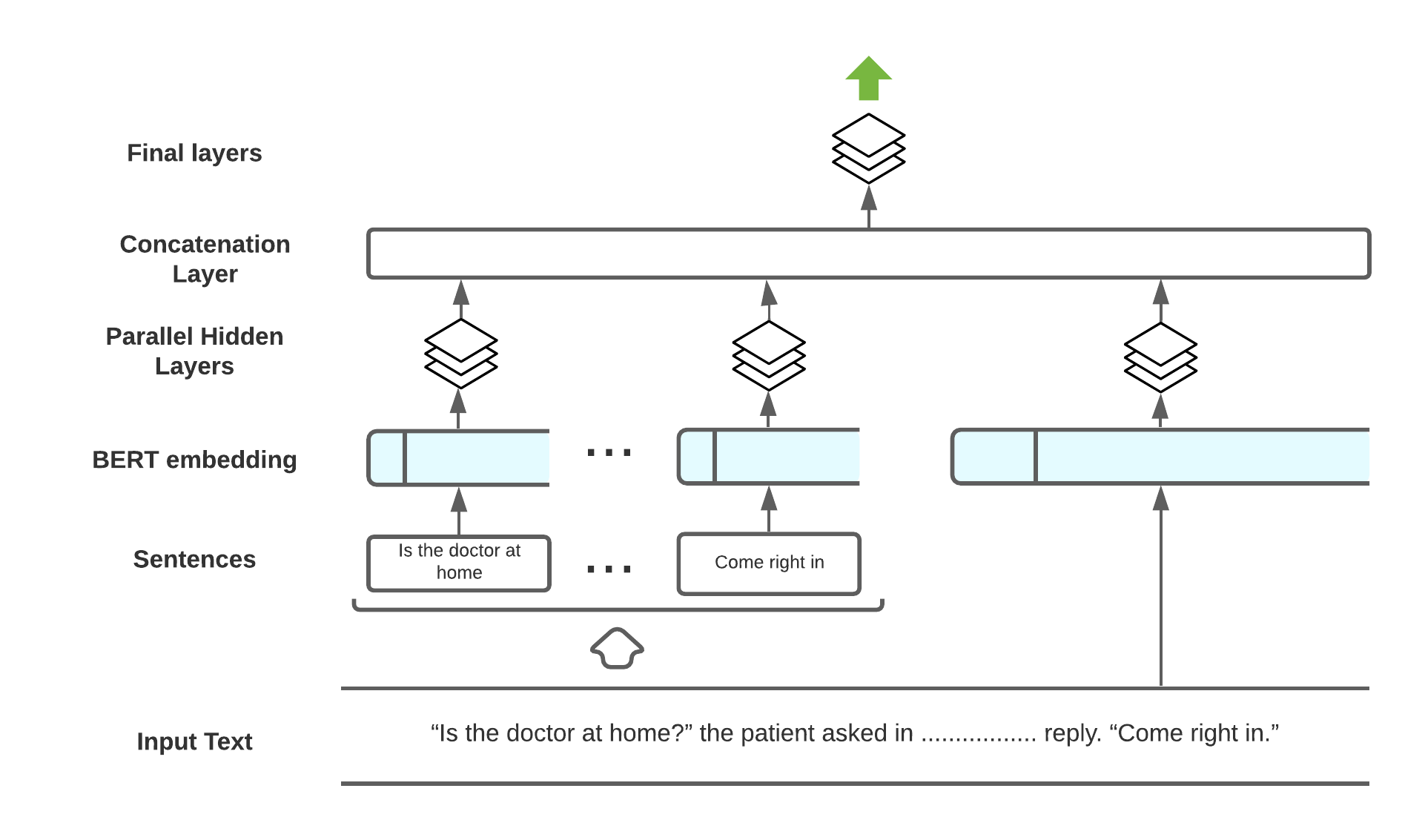}
 \caption{Components of the proposed method}
  \label{figure-1}
 \end {center}
\end{figure*}

Figure \ref{figure-1} displays the architecture of the proposed method. It contains separate paths of hidden layers specially designed to extract latent features from each sentence. Furthermore, there is a separate path to extract latent features of the whole text. Hence, our proposed neural network structure includes a single path to view the text as a whole and several other paths to view each sentence separately. It is comprised of a few general steps:

\begin{enumerate}
    \item First, to assess each sentence separately and extract numerical features, we separate sentences and tokenize them individually.
    \item To prepare these textual parts as proper numerical inputs for the neural network, we encode them using BERT sentence embedding. This step is performed individually on each sentence (left side in Figure \ref{figure-1}) and also on the whole text (right side in Figure \ref{figure-1}).
    \item Now that we have BERT sentence embedding for each sentence, we feed them into parallel hidden layers of neural network to extract mid-level features for each sentence (related to context, type of sentence, etc). The output of this part for each sentence is a vector of size 20.
    \item While our main idea is to detect existing relationships between sentences (specifically the punchline's relationship with the rest), it is also required to examine word-level connections in the whole text that may have meaningful impacts in determining congruity of the text. For example, existence of synonyms and antonyms in text could be meaningful. We feed BERT sentence embedding for the whole text into hidden layers of neural network (right side in Figure \ref{figure-1}). The output of this part is a vector of size 60.
    \item Finally, three sequential layers of neural network conclude our model. These final layers combine the output of all previous paths of hidden layers in order to predict the final output. In theory, these final layers should determine the congruity of sentences and detect the transformation of reader's viewpoint after reading the punchline.
\end{enumerate}

\subsection{Implementation Notes}
\label{sec:impl}
Since our approach builds on using BERT sentence embedding in a neural network, first, we obtain token representation using BERT tokenizer with the maximum sequence length of 100 (the maximum sequence length of BERT is 512). Then, we generate BERT sentence embedding by feeding tokens as input into the BERT model (vector size=768). 

The model will pass BERT embedding vectors of the given text and its sentences as inputs to a neural network with eight layers. For each sentence, We have a separate parallel line of three hidden layers which are concatenated in the fourth layer and continue in a sequential manner to predict the single target value. We use \verb+huggingface+ and \verb+keras.tensorflow+ packages for the BERT model and neural network implementations, respectively.

It is important to note that we used the BERT model to generate sentence embedding. Therefore, training is performed on the neural network and not on the BERT model. BERT comes with two pre-trained general types (the BERT\textsubscript{BASE} and the BERT\textsubscript{LARGE}), both of which are pre-trained from unlabeled data extracted from BooksCorpus \cite{zhu2015aligning} with 800M words and English Wikipedia with 2,500M words \cite{devlin2018bert}. In our proposed method, we use the smaller sized (BERT\textsubscript{BASE}) with 12 layers, 768-hidden states, 12-heads, and 110M parameters, which are pre-trained on lower-cased English text (uncased).

To achieve clean data, we performed a few textual preprocessing actions on all input texts. They are performed as part of the method and the novel dataset is not impacted:

\begin{itemize}
    \item \textbf{Expanding contractions}: We replaced all contractions with the expanded version of the expressions. For example, "is not" instead of “isn’t".
    \item \textbf{Cleaning punctuation marks}: We separated the punctuation marks\footnote{The punctuation marks are: period, comma, question mark, hyphen, dash, parentheses, apostrophe, ellipsis, quotation mark, colon, semicolon, exclamation point.} from words to achieve cleaner sentences. For example, the sentence “This is’ (fun).” is converted to “This is ‘ ( fun ) .”
    \item \textbf{Cleaning special characters}: We replaced some special characters with an alias. For example, “alpha” instead of “\( \alpha \)”.
\end{itemize}

\section{Evaluation and Discussion}

In this section, we evaluate the performance and robustness of the proposed method on two datasets. First, we compare the performance of the proposed model with a few baselines on the novel ColBERT dataset of formal English texts. Then, we report the performance evaluation on a dataset composed of informal tweets in the Spanish language.

\subsection{Evaluation on the ColBERT dataset}

The performance of the ColBERT model on the novel dataset is compared with five general baselines. In short, the dataset contains short formal humorous texts in English language. For the purposes of this section, the data is split into 80\% (160k) train and 20\% (40k) test parts.


We chose five strong baselines for comparison. The baseline models are:

\begin{enumerate}
    \item Decision Tree: A methodology that is commonly used as a data mining method for establishing classification systems based on multiple covariates or for developing prediction algorithms for a target variable. The method uses the train dataset to generate a branch-like segments that construct an inverted tree with a root node, internal nodes, and leaf nodes \cite{song2015decision}. For our evaluation, we used \verb+CountVectorizer+ to generate numerical word representations.
    \item SVM: A supervised model that achieved robust results for many classification and regression tasks. For this baseline, we applied \verb+TfidfVectorizer+ to generate numerical word representations with some optimization on hyper-parameters.
    \item Multinomial naïve Bayes: The model is suited when we deal with discrete integer features, such as word counts in a text. Here, we used \verb+CountVectorizer+ to generate numerical word representations.
    \item XGBoost: XGBoost is the latest step in the evolution of tree-based algorithms that include decision trees, boosting, random forests, boosting and gradient boosting. It is an optimized distributed gradient boosting that provides fast and accurate results, which achieves accurate results in less time \cite{chen2016xgboost}. We applied XGBoost on numerical word representations generated by \verb+CountVectorizer+ which resulted in better accuracy than \verb+TfidfVectorizer+.
    \item XLNet: A generalized language model that aims to mitigate the issues related to BERT model and previous autoregressive language models. For the task of text classification (and some other NLP tasks), XLNet outperforms BERT on several benchmark datasets \cite{yang2019XLNet}. We used \verb+xlnet-large-cased+ that has 24 layers and 340M parameters.
\end{enumerate}

We trained these baselines on using the cross-validation approach (K=5). Thus, in every fold, we used 128K for training of the model and the remaining 32K for evaluation. The following results are based on the final evaluation of the trained models on the test part of the dataset (remaining 40K).

\subsubsection{Results}

The results of our experiments on the ColBERT dataset are displayed in Table~\ref{table-3}. They found the proposed model’s accuracy and F1 score to be 98.2\%, thus outperforming all selected baselines with a large margin. This is a 7\% jump from the recent state-of-the-art XLNet model (with 340M parameters) and 17\% higher than the gradient boosting classifier. Traditional models of Decision Tree, SVM and Multinomial naïve Bayes gained less than 90\% accuracy in their, still an acceptable performance for a general model. XGBoost, a strong implementation of gradient boosting, achieved 81\% F1-score based on the selected word representations. XLNet \textsubscript{Large}, which required less optimization, was the strongest among the baselines, reaching close to 92\% accuracy, 4 percent higher than Multinomial NB.

Regarding time performance, the proposed model requires 2 hours (in average) to perform one epoch of training on 128k rows of the dataset on a computer with NVIDIA TESLA P100 GPUs. This is comparably less than the XLNet model, but longer than the rest of the selected baselines.

\begin{table*}[t]
  \caption{Performance evaluation on the ColBERT Dataset}
  \label{table-3}
  \centering
  \begin{tabular}{l|p{3.4cm}llll}
  \hline
Method        & Configuration                                                                              & Accuracy & Precision & Recall & F1    \\ \hline
Decision Tree &         & 0.786  & 0.769    & 0.821     & 0.794 \\ 
SVM           & sigmoid, gamma=1.0  & 0.872    & 0.869     & 0.880  & 0.874 \\
Multinomial NB &    alpha=0.2        & 0.876  & 0.863    & 0.902     & 0.882 \\
XGBoost &         & 0.720  & 0.753    & 0.777     & 0.813 \\
XLNet           & XLNet-Large-Cased  & 0.916    & 0.872     & 0.973  & 0.920 \\
    \hline
Proposed       &  & 0.982    & 0.990     & 0.974  & 0.982 \\ \hline
\end{tabular}
\end{table*}

\subsubsection{Discussion}

The results suggest two important factors in achieving high accuracy in the current task. First, methods that rely on pre-trained language models to produce sentence embedding outperform traditional methods of the word representation. XLNet and the proposed model both use their own embeddings and achieve much better results than other baselines, and the traditional methods of word representations such as TF-IDF could not break a limit even with the use of the latest classification boosting models (such as XGBoost). Second, our model with 110M parameters and 8 layers was able to outperform XLNet with 340M parameters and 24 layers, which could be a result of utilizing the linguistic structure of humor in designing the proposed model.

In addition, by reviewing the wrong predictions by our classifier, it is clear that the mislabeled items are generally close to (or even a part of) the items of the other class. For example, our model mislabeled the following two sentences as not humor: 

\begin{itemize}
\item "A recent study by UN has found Dexter to be the no 1 cause for ocean pollution"
\end{itemize}
\begin{itemize}
\item "One out of five dentists has the courage to speak their own mind"
\end{itemize}

On the contrast, the following news articles are mislabeled as humor:

\begin{itemize}
\item "If we treated men like we do women, would they cry more at work?" (News story: \footnote{https://www.huffpost.com/entry/if-we-treated-men-like-women-would-they-cry-more-at\_b\_5904e9ace4b084f59b49f99b})
\item "How do we keep alias generation off facebook? permanent mittens." (News story: \footnote{https://www.huffpost.com/entry/facebook-and-kids\_b\_1579207})
\end{itemize}

\subsection{Evaluation on informal Spanish tweets}

In the second part of the evaluation, we test the robustness of the proposed method by applying it to a new context. For this step, we participated in a recent shared task to detect and rate humor in Spanish tweets (HAHA 2021\footnote{https://www.fing.edu.uy/inco/grupos/pln/haha/}). In this way, we compete against real teams of machine learning engineers on a previously unseen dataset. The competition was organized as a part of the IberLEF 2021 forum and attracted seventeen active teams that competed via the CodaLab platform. The new setting is different from the previous evaluation method:

\begin{enumerate}
  \item The new task is to \textbf{detect and rate} humor in \textbf{Spanish}, which we have no linguistic knowledge of.
  \item The texts are \textbf{informal tweets}, differing in size and structure from the formal texts of the previous benchmark.
\end{enumerate}

To predict results, we did not change the model structure, hyper-parameters, or any of the pre-processing functions. However, in order to extract sentence embedding for Spanish texts, we changed the selected BERT model from the English BERT-base-uncased to a recent Spanish equivalent (BETO-uncased \cite{CaneteCFP2020}). This was a required and logical step to achieve meaningful embeddings.

The organizers provided a corpus of crowd-annotated tweets using a voting scheme with six options \cite{chiruzzo2020haha}: the tweet is not humorous, or the tweet is humorous and a score is given between one (not funny) to five (excellent). Data contains 36,000 tweets separated into training (24,000 tweets), development (6,000 tweets), and testing (6,000 tweets).


Based on the official results reported by the organizers \cite{haha2021overview}, the proposed model performed strongly and achieved the 2\textsuperscript{nd} place for rating humor with 0.6246 root mean square error (RMSE). In the binary task of humor detection, our model achieved the 3\textsuperscript{rd} place with 0.8696 F1 score. The results clearly indicate the robustness and stability of the proposed method in detecting humor in any given text. Full results and discussions for this evaluation are presented at \cite{annamoradnejad2021colbert}.

\subsection{Limitations and Future Work}
\label{sec:limits}

\textbf{Multimodal humor detection:} Some recent works focused on detecting humor in voice, videos, and pictures (e.g. \cite{hasan2019ur, choube2020punchline, patro2021multimodal, wu2021mumor}). While the focus of this work was on textual content, we believe that it is possible to apply the underlying idea of the proposed method to other types of media.

\textbf{Labeling accuracy:} We used an automated way to curate the accompanying dataset and classify them into two categories. As we saw in the evaluations, some news headlines can be considered humorous texts. While we evaluated our proposed method on a second dataset, the ColBERT dataset could be improved by human annotation to be better suited for the evaluation of future works.

\textbf{Other humor theories:} The proposed method is based on one popular theory of humor. There are other theories that discuss the cause of laughter in humans (See \cite{attardo2010linguistic, scheel2017definitions}). It would be interesting to create models based on alternative theories and compare the results of the execution, as a way to prove or quantify their accuracy. 

\textbf{Punchline detection and continuous texts:} The current work determines the existence of humor in a given text as a whole. While this has its own applications in the classification of short text posts and user commands, in many situations the text is continuous and there are no clear cut boundaries between the sentences. In those situations, it would become essential to pinpoint the parts of speech that contain humor or separate sentences based on context.

\section{Conclusion}
\label{sec:conclustion}

For ages, human beings have been fantasizing about humanoid robots indistinguishable from humans. In making that a reality, humor cannot be missed as a major human feature, which for its subjectivity, ambiguity, and semantic intricacies has been a difficult problem for researchers to tackle. This work contributes to this human fantasy and is paving the way for creating high-quality artificial intelligence systems (such as chatbots, virtual assistants, and even robots) injected with adjustable humor.

Our technical approach is based on injecting BERT sentence embedding into a neural network model that processes sentences separately in parallel lines of hidden layers. This conforms to a widely accepted theory of humor. Our method obtained F1 scores of 0.982 and 0.869 on two different settings and outperforms state-of-the-art models. Furthermore, we presented a novel dataset consisting of 200k formal short texts for the task of humor detection.

Based on our results, we identified two important factors in achieving high accuracy in the current task: 1) usage of sentence embeddings, and 2) utilizing the linguistic structure of humor in designing the proposed model. Results showed that our hypothesis on the structure of humor is valid and can be utilized to create very accurate systems of humor detection. Future work can adapt the proposed method for developing systems for other tasks of computational humor, or test the proposed parallel neural network in a wider range of text classification tasks.



\bibliographystyle{elsarticle-num} 
\bibliography{main}






\end{document}